\pdfoutput=1

\documentclass[11pt]{article}

\usepackage[final]{acl}

\usepackage{times}
\usepackage{latexsym}
\usepackage{graphicx}
\usepackage{soul}
\usepackage{url}
\usepackage{amsthm}
\usepackage{booktabs}
\usepackage{bm}
\usepackage{amsmath}
\usepackage{tablefootnote}

\usepackage{multirow}
\usepackage[ruled,linesnumbered]{algorithm2e}
\usepackage{longtable}
\usepackage{array}
\usepackage{subfigure}
\usepackage{stfloats}
\usepackage{multicol}
\usepackage{color}
\usepackage{booktabs} 
\usepackage{arydshln}
\usepackage{balance}
\usepackage{xspace}
\usepackage{enumitem}
\usepackage{amssymb}

\newcommand{\tabincell}[2]{\begin{tabular}{@{}#1@{}}#2\end{tabular}}

\newcommand{\hide}[1]{} 

\newcommand{\vpara}[1]{\vspace{1.5ex}\noindent\textbf{#1}}
\newcommand{\ipara}[1]{\vspace{0.03in}\noindent \textit{#1 }}

\newcommand{\beq}[1]{\vspace{-0.03in}\begin{equation}#1\end{equation}}

\newcommand{\beqn}[1]{\begin{eqnarray}#1\end{eqnarray}}

\newcommand{\model}{SR}
\newcommand{\smodel}{SR }

\usepackage[T1]{fontenc}

\usepackage[utf8]{inputenc}

\usepackage{microtype}

\title{Subgraph Retrieval Enhanced Model for Multi-hop Knowledge Base Question Answering}

\author{Jing Zhang\textsuperscript{1}, Xiaokang Zhang\textsuperscript{1}, Jifan Yu\textsuperscript{2}, Jian Tang\textsuperscript{3}, \\\textbf{Jie Tang}\textsuperscript{2}, \textbf{Cuiping Li}\textsuperscript{1}\thanks{       $\text{ }$ \scriptsize Corresponding author.\normalsize}, \textbf{Hong Chen}\textsuperscript{1} \\
  \textsuperscript{1}School of Information, Renmin University of China, Beijing, China, \\
  \textsuperscript{2}Department of Computer Science and Technology, Tsinghua University,  Beijing, China \\
  \textsuperscript{3}Mila - Quebec AI Institute.\\
  \texttt{\{zhang-jing,zhang2718,licuiping,chong\}@ruc.edu.cn},\\ \texttt{yujf18@mails.tsinghua.edu.cn}, \texttt{jian.tang@hec.ca},\\ \texttt{jietang@tsinghua.edu.cn}
  }

\begin{document}
\maketitle

\begin{abstract}

Recent works on knowledge base question answering (KBQA) retrieve subgraphs for easier reasoning. The desired subgraph is crucial as a small one may exclude the answer but a large one might introduce more noises.
However, the existing retrieval is either heuristic or interwoven with the reasoning, causing reasoning on the partial subgraphs, which increases the reasoning bias when the intermediate supervision is missing.
This paper proposes a trainable subgraph retriever (\model) decoupled from the subsequent reasoning process, which enables a plug-and-play framework to enhance any subgraph-oriented KBQA model. 
Extensive experiments demonstrate \smodel achieves significantly better retrieval and QA performance than existing retrieval methods. 
Via weakly supervised pre-training as well as the end-to-end fine-tuning, \smodel achieves new state-of-the-art performance when combined with NSM~\cite{He_2021}, a subgraph-oriented reasoner, for embedding-based KBQA methods. 
Codes and datasets are available online\footnote{\scriptsize  https://github.com/RUCKBReasoning/SubgraphRetrievalKBQA \normalsize}.
\end{abstract}

\section{Introduction}
\label{sec:introduction}

Knowledge Base Question Answering (KBQA)~\cite{ZHANG202114} aims to seek answers to factoid questions from structured KBs such as Freebase, Wikidata, and DBPedia. KBQA has attracted a lot of attention, as the logically organized entities and their relations are beneficial for inferring the answer. 
Semantic parsing-based (SP-based) methods~\cite{das2021case, lan-jiang-2020-query, sun2020sparqa} and embedding-based methods~\cite{He_2021,sun-etal-2018-open, SunACL2019} are two mainstream methods for addressing KBQA. The former ones heavily rely on the expensive annotation of the intermediate logic form such as SPARQL.
Instead of parsing the questions, the later ones directly represent and rank entities based on their relevance to input questions. Among them, the models which first retrieve a question-relevant subgraph and then perform reasoning on it~\cite{He_2021, sun-etal-2018-open, SunACL2019} reduce the reasoning space, showing superiority compared with reasoning on the whole KB~\cite{chen2019bidirectional, saxena-etal-2020-improving, xu-etal-2019-enhancing} (Cf. Table~\ref{tb:overall} for empirical proof).

\begin{figure}[t]
	\centering
	\subfigure[Answer Coverage Rate]{\label{subfig:coveragerate}
		\includegraphics[width=0.23\textwidth]{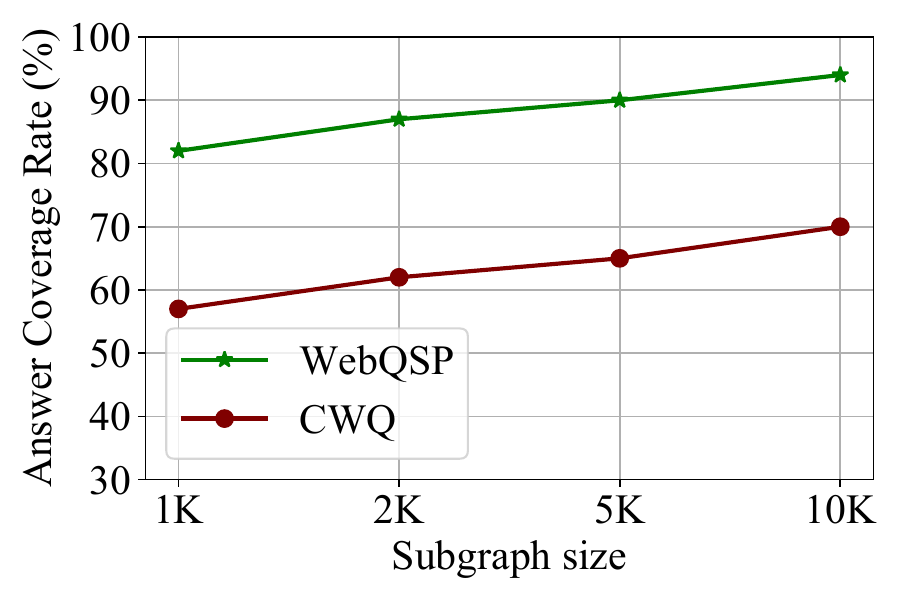}
	}
	\hspace{-0.1in}
	\subfigure[Hits@1 of QA]{\label{subfig:qa}
		\includegraphics[width=0.23\textwidth]{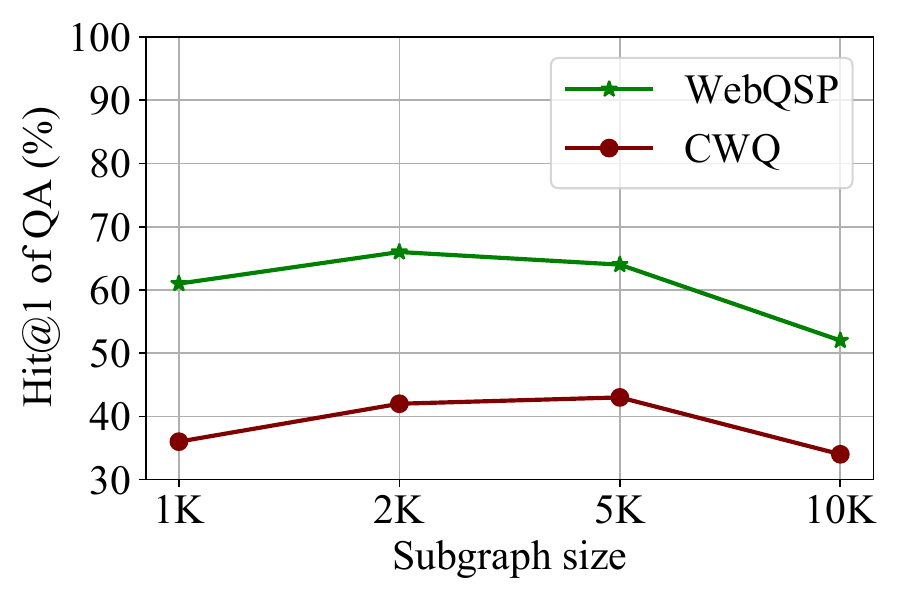}
	}
	\caption{\label{fig:graphsize} The impact of subgraph size on (a) answer coverage rate and (b) QA performance (Hits@1) of NSM~\cite{He_2021} on WebQSP~\cite{yih-etal-2016-value} and CWQ ~\cite{talmor-berant-2018-web}.}
\end{figure}

Subgraph retrieval is crucial to the overall QA performance, as a small subgraph is highly likely to exclude the answer but a large one might introduce noises that affect the QA performance.
Figure~\ref{subfig:coveragerate} presents the answer coverage rates of the subgraphs with different sizes on two widely-used KBQA datasets, WebQSP~\cite{yih-etal-2016-value} and CWQ~\cite{talmor-berant-2018-web}.
We extract the full multi-hop topic-centric subgraph and control the graph size by the personalized pagerank (PPR) ~\cite{haveliwala2003topic} scores of entities. We also present the QA performance (Hits@1) of NSM~\cite{He_2021}, a state-of-the-art embedding-based model, under the same sizes of the subgraphs in Figure~\ref{subfig:qa}.
It is observed that although larger subgraphs are more likely to cover the answer, the QA performance drops dramatically when the subgraph includes more than 5,000 nodes. 
Moreover, it is inefficient to extract such a full multi-hop subgraph for online QA. 
The results show that such heuristic retrieval is far from optimal.
To improve the retrieval performance, PullNet~\cite{SunACL2019} proposes a trainable retriever, but the retrieving and the reasoning processes are intertwined. 
At each step, a LSTM-based retriever selects new relations relevant to the question, and a GNN-based reasoner determines which tail entities of the new relations should be expanded into the subgraph. 
As a result, the inference as well as the training of the reasoner needs to be performed on the intermediate partial subgraph.
Since the intermediate supervision is usually unobserved, reasoning on partial subgraphs increases the bias which will eventually affect the answer reasoning on the final entire subgraph.


This paper proposes a subgraph retrieval enhanced model for KBQA, which devises a trainable subgraph retriever (\model) decoupled from the subsequent reasoner. \smodel is devised as an efficient dual-encoder that can expand paths to induce the subgraph and can stop the expansion automatically. After that, any subgraph-oriented reasoner such as GRAFT-Net~\cite{sun-etal-2018-open} or NSM~\cite{He_2021} can be used to delicately deduce the answers from the subgraph. 
Such separable retrieval and reasoning ensure the reasoning only on the final entire instead of the intermediate partial subgraphs, which enables a plug-and-play framework to enhance any subgraph-oriented reasoner.

We systematically investigate the advantages of various training strategies for \model, including weakly supervised/unsupervised pre-training and end-to-end fine-tuning with the reasoner.
Instead of the ground truth paths, we extract the shortest paths from a topic entity in the question to an answer as the weak supervision signals for pre-training. When the QA pairs themselves are also scarce, we construct pseudo (question, answer, path) labels for unsupervised pre-training.
To further teach the retriever by the final QA performance, we enable the end-to-end fine-tuning, which injects the likelihood of the answer conditioned on a subgraph as the feedback from the reasoner into the prior distribution of the subgraph to update the retriever.

We conduct extensive experiments on WebQSP and CWQ. The results reveal four major advantages: (1) \model, combined with existing subgraph-oriented reasoners, achieves several gains (+0.4-9.7\% Hits@1 and 1.3-8.7\% F1) over the same reasoner that is performed with other retrieval methods. Moreover, \smodel together with NSM creates new state-of-the-art results for embedding-based KBQA models.
(2) With the same coverage rate of the answers, \smodel can result in much smaller subgraphs that can deduce more accurate answers.
(3) The unsupervised pre-training can improve about 20\% Hits@1 when none of the weak supervision data is provided. 
(4) The end-to-end fine-tuning can enhance the performance of the retriever as well as the reasoner.

\vpara{Contributions}. (1) We propose a trainable \smodel decoupled from the subsequent reasoner to enable a plug-and-play framework for enhancing any subgraph-oriented reasoner.
(2) We devise \smodel by a simple yet effective dual-encoder, which achieves significantly better retrieval and QA results than the existing retrieval methods.
(3) NSM equipped with \model, via weakly supervised pre-training and end-to-end fine-tuning, achieves new SOTA performance for embedding-based KBQA methods.

\section{Related Work}

\vpara{KBQA} solutions can be categorized into SP-based and embedding-based methods. SP-based methods~\cite{Bao-etal-2016-constraint,berant-liang-2014-semantic,das2021case,lan-jiang-2020-query,liang-etal-2017-neural,QiuCIKM20,sun2020sparqa} parse a question into a 
logic form that can be executed against the KB. These methods need to annotate expensive logic forms as supervision or are limited to narrow domains with a few logical predicates. 
Embedding-based methods embed entities and rank them based on their relevance to the question, where the entities are extracted from the whole KB~\cite{miller-etal-2016-key,saxena-etal-2020-improving} or restricted in a subgraph~\cite{chen2019bidirectional,He_2021,sun-etal-2018-open,zhang2017variational}.   
They are more fault-tolerant but the whole KB or the ad-hoc retrieved subgraph includes many irrelevant entities. 
Some works such as PullNet~\cite{SunACL2019}, SRN~\cite{qiu2020stepwise}, IRN~\cite{zhou-etal-2018-interpretable}, and UHop~\cite{chen2019uhop} enhance the retrieval by training the retriever, but the retrieving and the reasoning are intertwined, causing the reasoning on partially retrieved subgraphs. 
Because of such coupled design, the reasoner in SRN, IRN, and UHop is degenerated into a simple MLP. On the contrary, thanks to the decoupled design, the reasoner can be complicated to support more complex reasoning. Other works propose more complicated reasoner for supporting the numerical reasoning in KBQA~\cite{feng2021}.

\vpara{Open-domain QA} (OpenQA) aims to answer questions based on a large number of documents. Most of the OpenQA models also consist of a retriever to identify the relevant documents and a reasoner to extract the answers from the documents. 
The retriever is devised as a sparse term-based method such as BM25~\cite{robertson2009probabilistic} or a trainable dense passage retrieval method~\cite{karpukhin-etal-2020-dense,sachan-etal-2021-end}, and the reasoner deals with each document individually~\cite{guu2020retrieval} or fuses all the documents together~\cite{izacardleveraging2021}.
Different from the documents in openQA, the subgraphs in KBQA can be only obtained by multi-hop retrieval and the reasoner should deal with the entire subgraph instead of each individual relation to find the answer. Although some openQA research proposes multi-hop document retrieval~\cite{asai2020learning}, the focus is the matching of the documents rather than the relations to the questions in KBQA. Thus the concrete solution for KBQA should be different from openQA.
\section{Problem Definition}
\label{sec:problem}

\noindent A \textbf{knowledge base} (KB) $G$ organizes the factual information as a set of triples, \emph{i.e.}, $G =\{(e,r,e')|e,e'\in E, r\in R\}$, where $E$ and $R$ denote the entity set and the relation set respectively. 
Given a factoid question $q$, KBQA is to figure out the answers $A_q$ to the question $q$ from the entity set $E$ of $G$. The entities mentioned in $q$ are topic entities denoted by $E_q=\{e_q\}$, which are assumed to be given.
This paper considers the complex
questions where the answer entities are multi-hops away from the topic entities, called multi-hop KBQA.

\vpara{Probabilistic Formalization of KBQA.}
Given a question $q$ and one of its answers $a \in A_q$, we formalize the KBQA problem as maximizing the probability distribution $p(a|G,q)$. 
Instead of directly reasoning on $G$, we retrieve a subgraph $\mathcal{G} \subseteq G$ and infer $a$ on $\mathcal{G}$. Since $\mathcal{G}$ is unknown, we treat it as a latent variable and rewrite $p(a|G,q)$ as:

\beqn{
\label{eq:likelihood}
p(a|G,q) &=& \sum_{ \mathcal{G}} p_{\phi}(a|q,\mathcal{G}) p_{\theta}(\mathcal{G}|q).
}

In the above equation, the target distribution $p(a|G,q)$ is jointly modeled by a subgraph retriever $p_{\theta}(\mathcal{G}|q)$ and an answer reasoner $p_{\phi}(a|q,\mathcal{G})$. The subgraph retriever $p_{\theta}$ defines a prior distribution over a latent subgraph $\mathcal{G}$ conditioned on a question $q$, while the answer reasoner $p_{\phi}$ predicts the likelihood of the answer $a$ given $\mathcal{G}$ and $q$. The goal is to find the optimal parameters $\theta$ and $\phi$ that can maximize the log-likelihood of training data, \emph{i.e.}, 

\beq{
\small
\label{eq:objecitve}
    \mathcal{L}(\theta, \phi) = \max_{\theta, \phi} \sum_{(q,a,G) \in \mathcal{D}}\log \sum_{ \mathcal{G}} p_{\phi}(a|q,\mathcal{G}) p_{\theta}(\mathcal{G}|q),
}
\normalsize

\noindent where $\mathcal{D}$ is the whole training data. 
Thanks to this formulation, the retriever can be decoupled from the reasoner by firstly training the retriever $p_{\theta}$ and then the reasoner $p_{\phi}$ on the subgraphs sampled by the retriever. Via drawing a sample $\mathcal{G}$~\cite{sachan-etal-2021-end}, we can approximate Eq.~\eqref{eq:objecitve} as: 

\small
\beq{
\label{eq:approximateobjective}
\mathcal{L}(\theta, \phi) = \max_{\theta, \phi} \!\!\sum_{(q,a,\mathcal{G})\in \mathcal{D}}  \!\!\log p_{\phi}(a|q,\mathcal{G}) + \log p_{\theta}(\mathcal{G}|q),
} 
\normalsize

\noindent where the first and the second term can be optimized for the reasoner and the retriever respectively.
The concrete reasoner can be instantiated by any subgraph-oriented KBQA model such as the GNN-based GRAT-Net~\cite{sun-etal-2018-open} and NSM~\cite{He_2021}.

  \begin{figure*}[t]
	\centering
	\includegraphics[width=\textwidth]{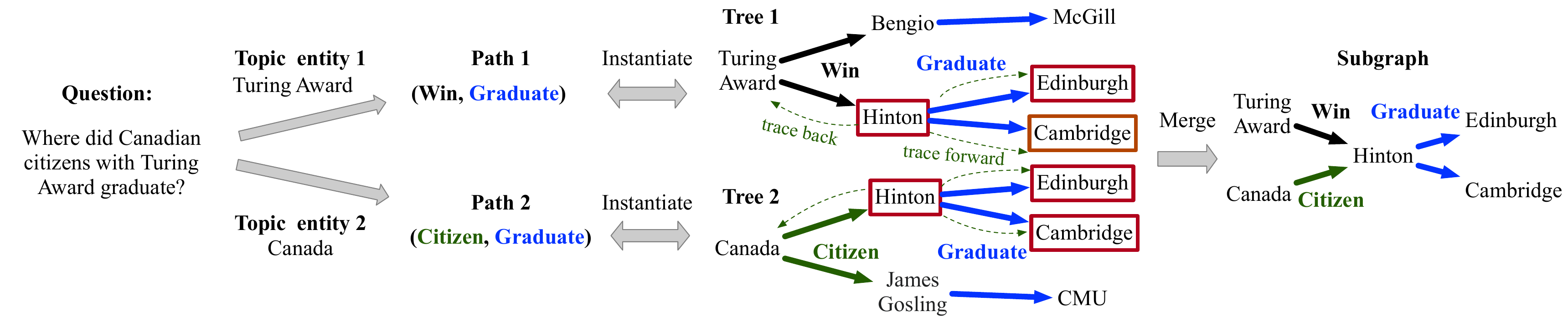}
	\caption{\label{fig:subgraph} Illustration of the subgraph retrieving process. We expand a path from each topic entity as well as induce a corresponding tree, and then merge the trees from different topic entities to form a unified subgraph.}
\end{figure*}
\section{Subgraph Retriever (\model)}
\label{sec:retriever}

The retriever needs to calculate $p_{\theta}(\mathcal{G}|q)$ for any $\mathcal{G}$, which is intractable as the latent variable $\mathcal{G}$ is combinatorial in nature.
To avoid enumerating $\mathcal{G}$, 
we propose to expand top-$K$ paths relevant to $q$ from the topic entities and then induce the subgraph following these paths. 

\subsection{Expanding Paths}
Path expanding starts from a topic entity and follows a sequential decision process. Here a path is defined as a sequence of relations $(r_1,\cdots, r_{|p|})$, since a question usually implies the intermediate relations excluding the entities. 
Suppose a partial path $p^{(t)} = (r_1, \cdots, r_{t})$ has been retrieved at time $t$, a tree can be induced from $p^{(t)}$ by filling in the intermediate entities along the path, \emph{i.e.}, $T^{(t)} = (e_q,r_1, E_1,\cdots, r_{t},E_{t})$. Each $E_t$ is an entity set as a head entity and a relation can usually derive multiple tail entities. Then we select the next relation from the union of the neighboring relations of $E_{t}$. The relevance of each relation $r$ to the question $q$ is measured by the dot product between their embeddings, \emph{i.e.},

\beq{
    s(q,r) = f(q)^{\top} h(r),
}

\noindent where both $f$ and $h$ are instantiated by RoBERTa~\cite{liu2019roberta}. 
Specifically, we input the question or the name of $r$ into RoBERTa and take its [CLS] token as the output embedding. 
According to the assumption~\cite{chen2019uhop,He_2021,qiu2020stepwise,zhou-etal-2018-interpretable} that expanding relations at different time steps should attend to specific parts of a query, we update the embedding of the question by simply concatenating the original question with the historical expanded relations in $p^{(t)}$ as the input of RoBERTa, \emph{i.e.},
\beq{
\label{eq:updatequestion}
    f(q^{(t)}) = \mbox{RoBERTa}([q;r_{1};\cdots;r_{t}]),
}

Thus $s(q,r)$ is changed to $s(q^{(t)},r)=f(q^{(t)})^{\top} h(r)$.
Then the probability of a relation $r$ being expanded can be formalized as:
\beq{
\small
\label{eq:tripletprobability}
    p(r|q^{(t)}) = \frac{1}{1 + \exp  \left(s(q^{(t)},\text{END}) - s(q^{(t)},r)\right)},
}
\normalsize

\noindent where $\text{END}$ is a virtual relation named as ``END''. The score $s(q^{(t)}, \text{END})$ represents the threshold of the relevance score. $p(r|q^{(t)})$ is larger than 0.5 if $s(q^{(t)},r) > s(q^{(t)},\text{END})$ and is no larger than 0.5 otherwise. 
We select the top-1 relation with $p(r|q^{(t)})>0.5$. The expansion is stopped if none of the probabilities of the relations is larger than 0.5.
Finally, the probability of a path given the question can be computed as the joint distribution of all the relations in the path, \emph{i.e.}, 

\beq{
\label{eq:pathprobability}
 p_{\theta}(p|q)= \prod_{t=1}^{|p|} p(r_t|q^{(t)}).
}

\noindent where $|p|$ denotes the number of relations in $p$, $t=1$ indicates the selection at the topic entity and $t=|p|$ denotes the last none-stop relation selection. Since the top-1 relevant path cannot be guaranteed to be right, we perform a top-$K$ beam search at each time to get $K$ paths. From each topic entity, we obtain $K$ paths which result in $nK$ paths in total by $n$ topic entities. $nK$ paths correspond to $nK$ instantiated trees.

\subsection{Inducing Subgraph}
We take the union of top-$K$ trees from one topic entity into a single subgraph, and then merge the same entities from different subgraphs to induce the final subgraph.
This can reduce the subgraph size, \emph{i.e.}, the answer reasoning space, as the subgraphs from different topic entities can be viewed as the constraints of each other. Specifically, from the $n$ subgraphs of the $n$ topic entities, we find the same entities and merge them. From these merged entities, we trace back in each subgraph to the root (\emph{i.e.}, a topic entity) and trace forward to the leaves. Then we only keep the entities and relations along the tracing paths of all the trees to form the final subgraph. For example in Figure~\ref{fig:subgraph}, given a question ``Where did Canadian citizens with Turing Award graduate?'' with two topic entities ``Turing Award'' and ``Canada''\footnote{Some work views ``Canada'' as a constraint, which is not easy to be distinguished with the topic entity ``Turing Award''. Thus this paper treats both of them as topic entities.}, we can explain it by the two expanded paths (Win, Graduate) and (Citizen, Graduate) and merge the trees induced by them to form a unified subgraph. Only the top-1 path is presented in the figure for a clear illustration.

 \begin{figure*}[t]
	\centering
	\includegraphics[width=0.9\textwidth]{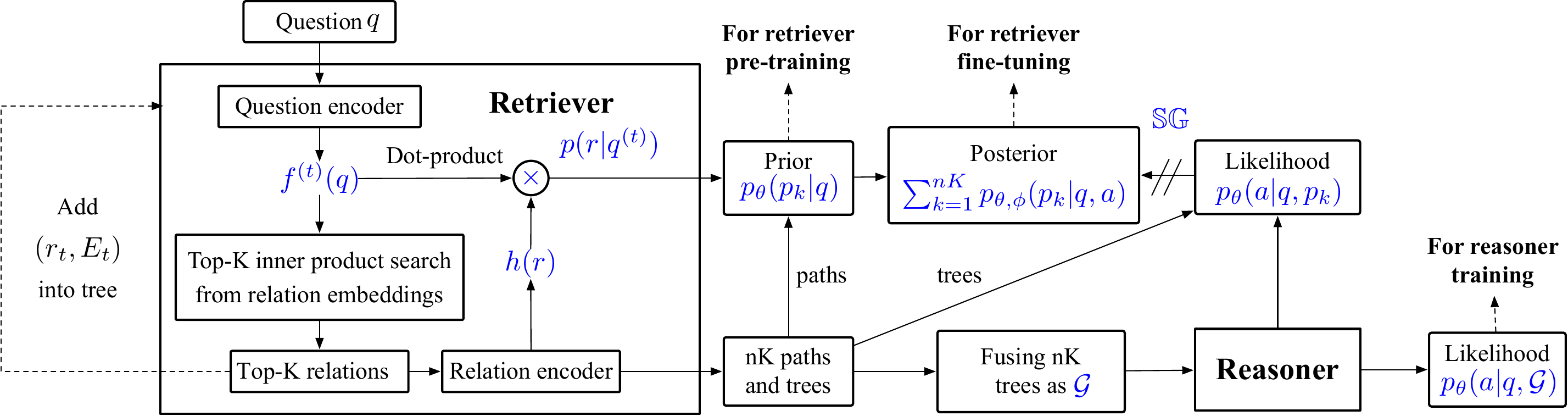}
	\caption{\label{fig:framework} Overview of \smodel and its training strategies. Given a question, \smodel generates $nK$ paths via iteratively expanding the relations. We pre-train the retriever based on the prior of each path and train the reasoner based on the likelihood of the subgraph fused from the $nK$ trees. For end-to-end training, the retriever is fine-tuned on the posterior of each path that consists of the prior and the likelihood of it. $\mathbb{SG}$ is the stop-gradient operation.}
\end{figure*}

\section{Training Strategies}
\label{sec:training}
In this section, we discuss the pre-training and the end-to-end fine-tuning strategies to train the retriever. 
Figure~\ref{fig:framework} illustrates the whole framework and the training procedure. 

\subsection{Weakly Supervised Pre-Training}
Since the ground truth subgraphs are not easy to be obtained, we resort to the weak supervision signals constructed from the $(q, a)$ pairs.
Specifically, from each topic entity of a question, we retrieve all the shortest paths to each answer as the supervision signals, as paths are easier to be obtained than graphs. 
Since maximizing the log-likelihood of a path equals to $\sum_{t=1}^{|p|} \log p(r_{t}|q^{(t)})$ according to Eq.~\eqref{eq:pathprobability}, 
we can maximize the probabilities of all the intermediate relations in a path.
To achieve the goal, we decompose a path $p = (r_1,\cdots, r_{|p|})$ into $|p|+1$ (question, relation) instances, including $([q],r_1)$, $([q;r_1], r_2)$, ..., $([q;r_1;r_2;\cdots; r_{|p|-1}],r_{|p|})$, and an additional END instance $([q;r_1;r_2;\cdots; r_{|p|}],\text{END})$, and optimize the probability of each instance.
We replace the observed relation at each time step with other sampled relations as the negative instances to optimize the probability of the observed ones.

\subsection{Unsupervised Pre-Training}
When the $(q, a)$ pairs are also scarce, we train the retriever in an unsupervised manner independent from the $(q, a)$ pairs. 
We leverage the NYT dataset, a distant supervision dataset for relation extraction ~\cite{riedel2010modeling} to construct the pseudo $(q,a,p)$ labels. In this dataset, each instance is denoted as a tuple $(s, (e_1,r,e_2))$, where $s$ is a sentence that refers to the relation $r$ between two entities $e_1$ and $e_2$ mentioned in the sentence $s$. For two instances $(s_1,(e_1,r_1,e_2))$ and $(s_2,(e_2,r_2,e_3))$, we treat $e_1$ as the topic entity and $e_3$ as the answer. Then we concatenate $s_1$ and $s_2$ as the question, and concatenate $r_1$ and $r_2$ as the corresponding path to train the retriever. The training objective is the same as the weakly supervised pre-training.

\subsection{End-to-End Fine-tuning}
End-to-end training is an alternative to fine-tune the separately trained retriever and the reasoner jointly.
The main idea is to leverage the feedback from the reasoner to guide the path expansion of the retriever. 
To enable this, we optimize the posterior $p_{\theta,\phi}(\mathcal{G}|q,a)$ instead of the prior $p_{\theta}(\mathcal{G}|q)$, since the former one contains the additional likelihood $p_{\phi}(a|q,p_k)$ which exactly reflects the feedback from the reasoner.
We do not directly optimize the posterior $p_{\theta,\phi}(\mathcal{G}|q,a)$, because $\mathcal{G}$ is induced from $nK$ paths, making it unknown which path should receive the feedback from the likelihood computed on the whole $\mathcal{G}$.
Instead, we approximate $p(\mathcal{G}|q,a)$ by the sum of the probabilities of the $nK$ paths and rewrite the posterior of each path by Bayes' rule~\cite{sachan-etal-2021-end}, \emph{i.e.},
\beqn{
    p_{\theta,\phi}(\mathcal{G}|q,a) &\approx& \sum_{k=1}^{nK} p_{\theta,\phi}(p_k |q, a), \\
    &\propto& \sum_{k=1}^{nK} p_{\phi}(a|q,p_k)p_{\theta}(p_k|q), \nonumber
 }

\noindent where $p_{\theta}(p_k|q)$ is the prior distribution of the $k$-th path that can be estimated by Eq.~\eqref{eq:pathprobability}, and $p_{\phi}(a|q,p_k)$ is the likelihood of the answer $a$ given the $k$-th path. Essentially, $p_{\phi}(a|q,p_k)$ estimates the answer $a$ on the single tree induced by the $k$-th path instead of the fused subgraph by $nK$ paths. As a result, the reasoning likelihood on each tree can be reflected to the corresponding path that induces the tree.
The reasoner for estimating $p_{\phi}(a|q,p_k)$ is the same as that for calculating $p_{\phi}(a|q,\mathcal{G})$.

In summary, the whole objective function for each training instance $(q,a,\mathcal{G})$ is formalized as:
\beqn{
\label{eq:loss}
\mathcal{L} &=&  \underbrace{\max_{\phi} \log p_{\phi}(a|q,\mathcal{G})}_{\text{Reasoner}} \\ \nonumber
&+& 
\underbrace{\max_{\theta} \log \sum_{k=1}^{nK} \mathbb{SG}(p_{\phi}(a|q,p_k))p_{\theta}(p_k|q)}_{\text{Retriever}},
}

\noindent where the stop-gradient operation $\mathbb{SG}$ is to stop updating the parameters $\phi$. The reasoner is updated the same as the two-stage training by computing the likelihood $p_{\phi}(a|q,\mathcal{G})$ on $\mathcal{G}$ sampled by the retriever (without using information from the answer $a$). As a result, there is no mismatch between the training and evaluation when computing $p_{\phi}(a|q,\mathcal{G})$, as $\mathcal{G}$ relies only on the prior at both.

Intuitively, we train the reasoner to extract the correct answer given the subgraph induced from $nK$ highest scoring paths. And we train the retriever to select $nK$ paths which collectively have a high score to deduce the answer when taking the feedback from the reasoner into account. Although the two components are jointly trained, the reasoning is still performed on the retrieved entire subgraph at each epoch. We present the training process in Appendix.

\section{Experiments}
\label{sec:exp}

In this section, we conduct extensive experiments to evaluate the subgraph retrieval (\model) enhanced model.
We design the experiments to mainly answer the four questions: 
(1) Does \smodel take effect in improving the QA performance?
(2) Can \smodel obtain smaller but higher-quality subgraphs?
(3) How does the weakly supervised and unsupervised pre-training affect \model's performance?
(4) Can end-to-end fine-tuning enhance the performance of the retriever as well as the reasoner?

\subsection{Experimental Settings}
\vpara{Datasets.}
We adopt two benchmarks, WebQuestionSP (WebQSP)~\cite{yih-etal-2016-value} and Complex WebQuestion 1.1 (CWQ)~\cite{talmor-berant-2018-web}, for evaluating the proposed KBQA model.  
Table~\ref{tb:dataset} shows the statistics.

\vpara{Evaluation Metrics.} 
We evaluate the retriever by the answer coverage rate, which is the proportion of questions for which the topic-$nK$ retrieved paths contain at least one answer. This metric reflects the upper bound of the QA performance and is denoted as Hits@$K$.
For QA performance, We use Hits@1 to evaluate whether the top-1 predicted answer is correct. Since some questions have multiple answers, we also predict the answers by the optimal threshold searched on the validation set and evaluate their F1 score.

\begin{table}[t]
\newcolumntype{?}{!{\vrule width 1pt}}
\newcolumntype{C}{>{\centering\arraybackslash}p{2em}}
\caption{
	\label{tb:dataset} Data statistics. The number of QA pairs for training, validating and testing are presented.  
}
\centering 
\renewcommand\arraystretch{1.0}
\scalebox{0.9}{
\begin{tabular}{@{}l?ccc@{}}
\toprule
Dataset & \#Train       & \#Validation        & \#Test      \\ \midrule
WebQSP   & 2,848     & 250      & 1,639  \\
CWQ      & 27,639  & 3,519   & 3,531 \\ \bottomrule
\end{tabular}
}
\end{table}

\vpara{Baseline Models.}
We compare with embedding-based KBQA models, in which
EmbedKGQA~\cite{saxena-etal-2020-improving} directly optimizes the triplet (topic entity, question, answer) based on their direct embeddings. KV-Mem~\cite{miller-etal-2016-key}  BAMNet~\cite{chen2019bidirectional} store triplets in a key-value structured memory for reasoning.
GRAFT-Net~\cite{sun-etal-2018-open}, BAMNet~\cite{chen2019bidirectional}, NSM~\cite{He_2021}, and PullNet~\cite{SunACL2019} 
are subgraph-oriented embedding models.
We also compare with the SP-based models, 
in which QGG~\cite{lan-jiang-2020-query} generates the query graph for a question by adding the constraints and extending the relation paths simultaneously, SPARQA~\cite{sun2020sparqa} proposes a novel skeleton grammar to represent a question, and CBR-KBQA~\cite{das2021case} leverages BigBird~\cite{zaheer2020big}, a pre-trained seq2seq model to directly parse a question into a SPARQL statement that can be executed on graph DBs. \smodel is default trained by weakly supervised pre-training and the default path number is set to 10.

\begin{table}[t]
\newcolumntype{?}{!{\vrule width 1pt}}
	\newcolumntype{C}{>{\centering\arraybackslash}p{2em}}
	\caption{
		\label{tb:overall} QA performance on WebQSP and CWQ (\%).
	}
	\centering 
	\renewcommand\arraystretch{1.0}
\scalebox{0.9}{
\begin{tabular}{@{}l?cc?cc}
\toprule
\multirow{2}{*}{Model} & \multicolumn{2}{c?}{WebQSP} & \multicolumn{2}{c}{CWQ}  \\  & Hits@1 & F1 &  Hits@1 & F1  \\\midrule
\multicolumn{5}{c}{SP-based models} \\\midrule
SPARQA  & - & - & 31.6 & - \\
QGG  & 73.0 & 73.8 & 36.9 &  37.4\\
CBR-KBQA  & - & 72.8 & - & \textbf{70.0} \\
\midrule
\multicolumn{5}{c}{Embedding-based models} \\\midrule
KV-Mem &46.6 & 34.5 &18.4 & 15.7\\
EmbedKGQA   & 66.6  & -& 32.0 & -\\
BAMnet & 55.6 & 51.8 & - & - \\
GRAFT-Net (GN)   & 66.4  & 60.4 & 36.8  & 32.7 \\ 
NSM         & 68.5  &  62.8   & 46.3  &  42.4  \\
PullNet     & 68.1  &  -   & 45.9  &  - \\

\midrule
\multicolumn{5}{c}{Our Models}\\
\midrule
\model+NSM & 68.9 & 64.1 & \textbf{50.2} & \textbf{47.1}\\      
\model+GN  & 65.2 & 61.2 & 46.5 & 41.4   \\    
\model+NSM w E2E & \textbf{69.5} & \textbf{64.1} & 49.3 & 46.3 \\     
\model+GN w E2E  & 66.7 & 63.1 & 49.0 & 42.7\\  
\bottomrule
\end{tabular}
}
\end{table}

\subsection{Overall QA Evaluation}
We compare with state-of-the-art KBQA models and present the Hits@1 and F1 scores in Table~\ref{tb:overall}.

\vpara{SP-based Models.}
The SP-based model CBR-KBQA achieves the best performance on CWQ. This is expected, as CBR-KBQA leverages a pre-trained seq-to-seq model to parse the input question into a SPARQL statement. However, the model depends on the annotated SPARQL statements, which are expensive to be annotated in practice.

\vpara{Embedding-based Models.}
Among these models, KV-Mem and EmbedKGQA retrieve the answers from the global key-value memory built on the KB or the original whole KB, which enjoys high recall but suffers from many noisy entities. 
Compared with these global retrievals, 
BAMNet builds the key-value memory on a  subgraph, but it is a full multi-hop topic-entity-centric subgraph, which is also noisy.
GRAFT-Net and NSM calculate PPR scores to control the subgraph size, but the ad-hoc retrieval method is still far from optimal. PullNet reinforces the retrieval by learning a retriever, but the retriever and the reasoner are intertwined, causing the partial reasoning on part of a subgraph, which increases the reasoning bias.

\begin{table*}[t]
\newcolumntype{?}{!{\vrule width 1pt}}
	\newcolumntype{C}{>{\centering\arraybackslash}p{2em}}
	\caption{
		\label{tb:retrieverperformance} 
		The answer coverage rate of \smodel on WebQSP and CWQ (\%). 
	}
	\centering 
	\renewcommand\arraystretch{1.0}
\scalebox{0.9}{
\begin{tabular}{@{}l@{ }?cccc?cccc@{}}
\toprule
\multirow{2}{*}{Model} & \multicolumn{4}{c?}{WebQSP} & \multicolumn{4}{c}{CWQ}  \\  & Hits@1 & Hits@5 & Hits@10 & Hits@20 & Hits@1 & Hits@5 & Hits@10 & Hits@20 \\
\midrule
\smodel w SuperT  & 78.0 &  90.7  & 90.5 & 95.0 & -    & -    & -    & - \\
\model            & 71.2 &  87.5  & 90.5 & 92.9 & 51.5 & 72.6 & 79.5 & 83.7 \\
\smodel w/o QU    & 66.9 &  82.6  & 86.2 & 89.4 & 36.5 & 46.4 & 50.8 & 54.9\\
\smodel w/o PE    & 52.7 &  71.4  & 74.8 & 78.1 & 49.4 & 62.3 & 66.6 & 71.6\\
\model+NSM w E2E  & 79.3 &  89.8  & 91.6 & 93.3 & 53.5 & 73.4 & 78.4 & 82.8  \\
\model+GN w E2E   & 81.8 &  90.2  & 92.9 & 94.4 & 56.1 & 74.5 & 79.6 & 83.9 \\

\bottomrule
\end{tabular}
}
\end{table*}

\vpara{Our Models.}
Compared with the above embedding-based models, a performance improvement on both the datasets can be observed, \emph{e.g.}, NSM injected by \smodel (\model+NSM) improves 0.4\% Hits@1 and 1.3\% F1 on WebQSP, 3.9\% Hits@1 and 4.7\% F1 on CWQ compared with the original NSM.
We also show that \smodel can be adapted to different subgraph-oriented reasoners. Beyond NSM, when injecting \smodel to GRAFT-NET, it also significantly improves 9.7\% Hits@1 and 8.7\%F1 on CWQ. \model+GN underperforms GN on WebQSP because GN filters out the relations of the knowledge graph not in the training set of WebQSP. We do not inject \smodel into BAMNet as the model needs entity types in the subgraph, which is temporarily ignored by \smodel. 

\ipara{Summary.} The overall evaluation shows that \smodel takes effect in improving the QA performance when injecting it before a subgraph-oriented reasoner, and \smodel equipped with NSM creates a new state-of-the-art model for embedding-based KBQA.

\begin{figure}[t]
	\centering
	\subfigure[WebQSP]{\label{subfig:webqspszvsacr}
		\includegraphics[width=0.23\textwidth]{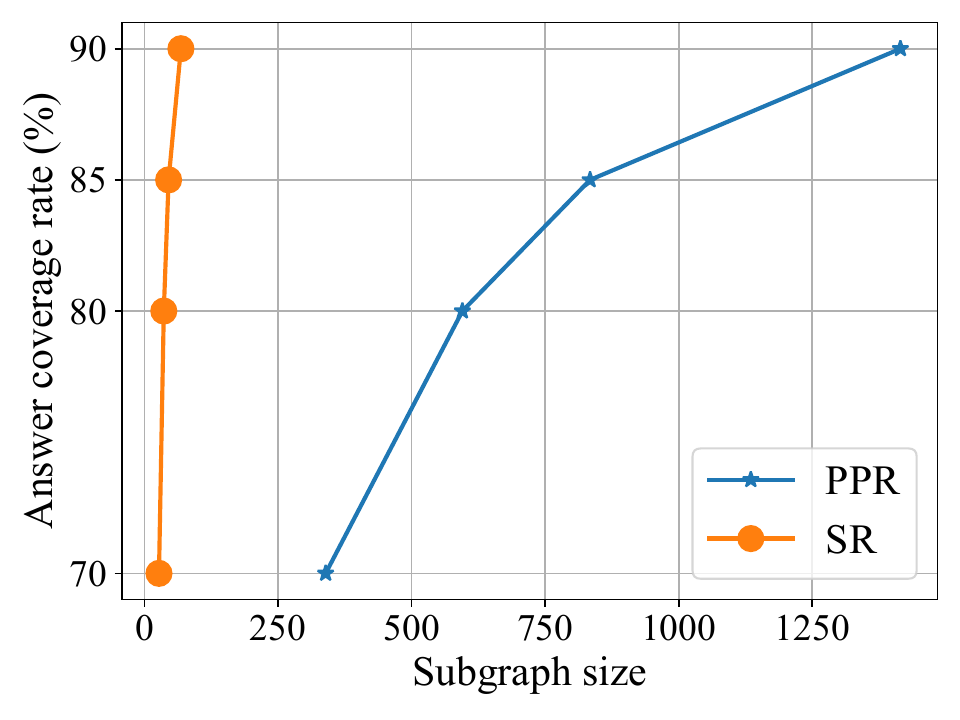}
	}
	\hspace{-0.1in}
    \subfigure[CWQ]{\label{subfig:cwqszvsacr}
		\includegraphics[width=0.23\textwidth]{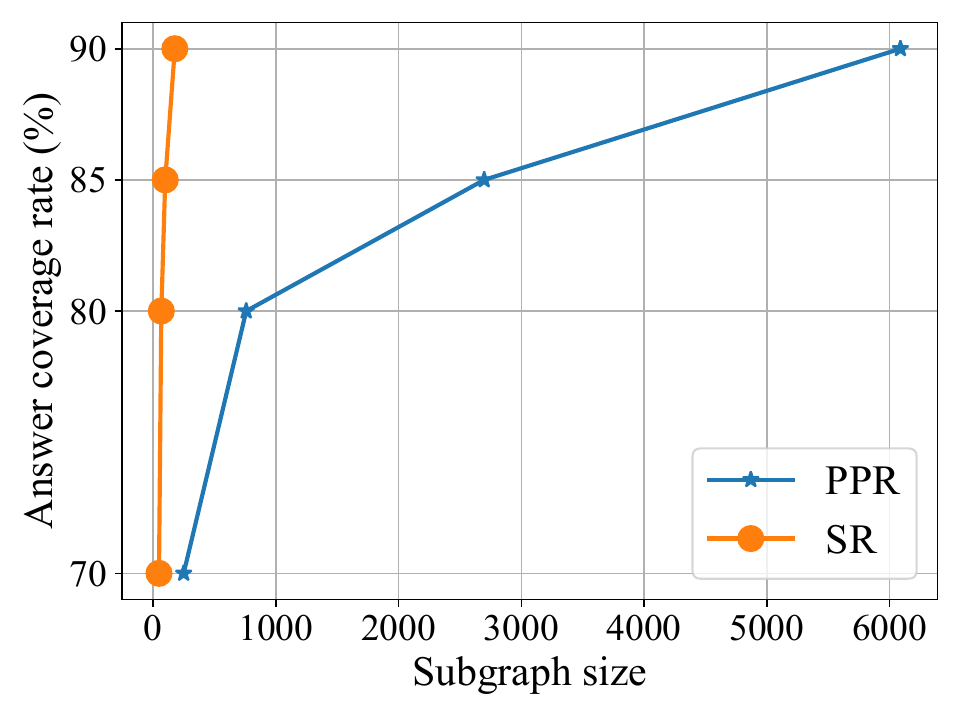}
	}
	\hspace{-0.1in}
    \subfigure[WebQSP]{\label{subfig:webqspacrvsqa}
		\includegraphics[width=0.23\textwidth]{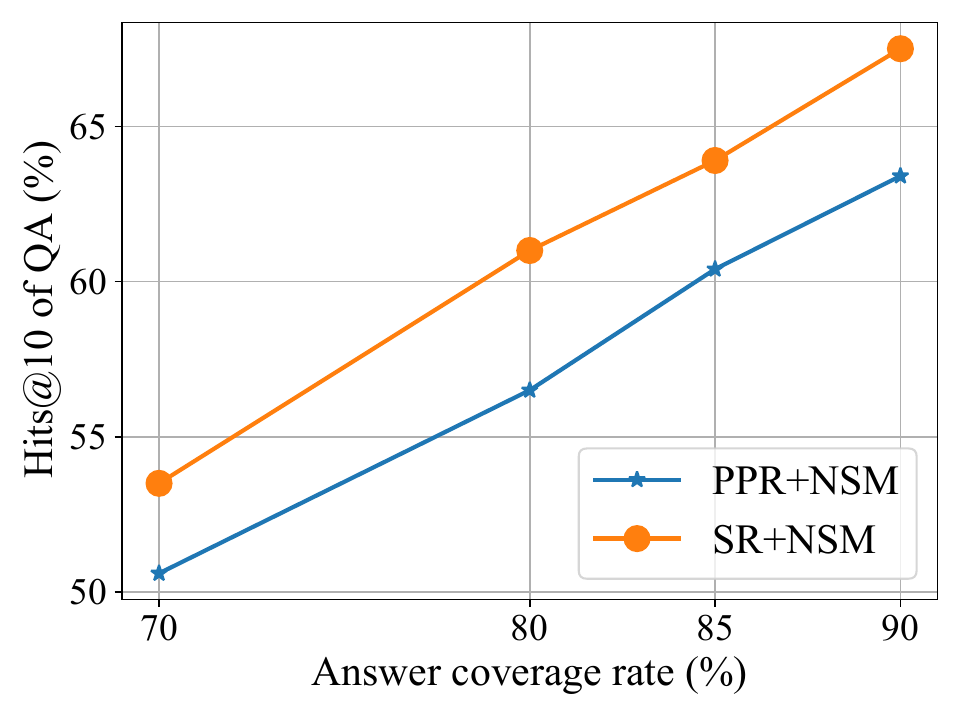}
	}
	\hspace{-0.1in}
    \subfigure[CWQ]{\label{subfig:cwqacrvsqa}
		\includegraphics[width=0.23\textwidth]{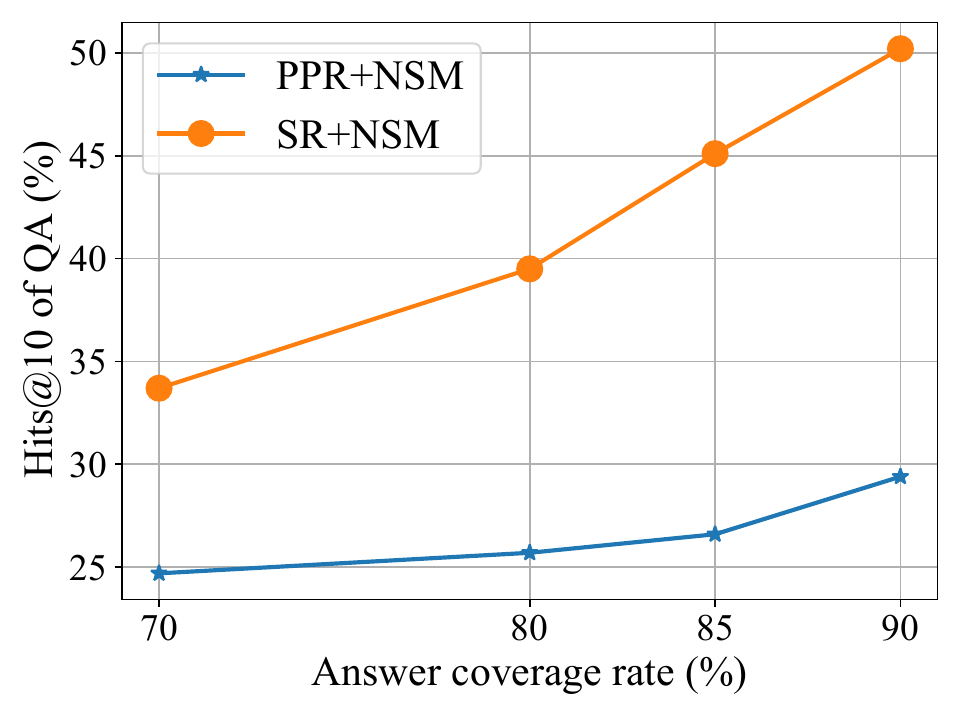}
	}
	
	\caption{\label{fig:subgraphsizeanswercoveragerate} Comparison of the answer coverage rate under various subgraph sizes (Top row) and the QA performance (Hits@1) under various answer coverage rates (Bottom row).
	}
\end{figure}

\subsection{Retriever Evaluation}

\vpara{Quality of Retrieved Subgraph.}
We evaluate whether the proposed \smodel can obtain smaller but higher-quality subgraphs, which are measured by not only the direct subgraph size and answer coverage rate but also the final QA performance.
For a fair comparison, we fix the reasoner as NSM, and vary the retriever as \smodel and the PPR-based heuristic retrieval~\cite{sun-etal-2018-open,He_2021}. PPR+NSM are performed on the same knowledge graph of the proposed \model+NSM. 
The result of the trainable retriever in PullNet~\cite{SunACL2019} is ignored, because its code is not published and the value of some key parameters that seriously impact the model's performance is unknown.

We report the comparison results in Figure~\ref{fig:subgraphsizeanswercoveragerate}. The top row presents the answer coverage rates of the subgraphs with various sizes. It is shown that when retrieving the subgraphs of the same size, the answer coverage rate of \smodel is significantly higher than PPR. The bottom row presents the QA performance (Hits@1) on the subgraphs with various answer coverage rates. It is shown that by performing the same NSM on the subgraphs with the same coverage rate, the subgraphs retrieved by \smodel can result in higher QA performance than PPR.

\ipara{Summary.} The above results show that \smodel can obtain smaller but higher-quality subgraphs.

\begin{figure}[t]
	\centering
	\subfigure[WebQSP]{\label{subfig:webqspsubgraphsize}
		\includegraphics[width=0.23\textwidth]{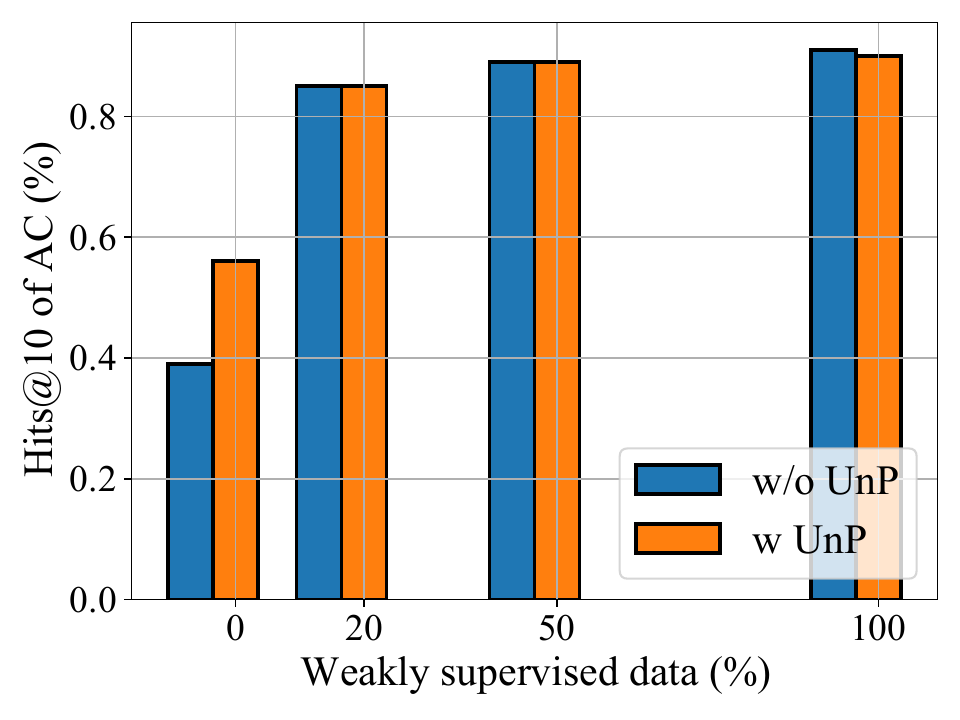}
	}
	\hspace{-0.1in}
    \subfigure[CWQ]{\label{subfig:cwqsubgraphsize}
		\includegraphics[width=0.23\textwidth]{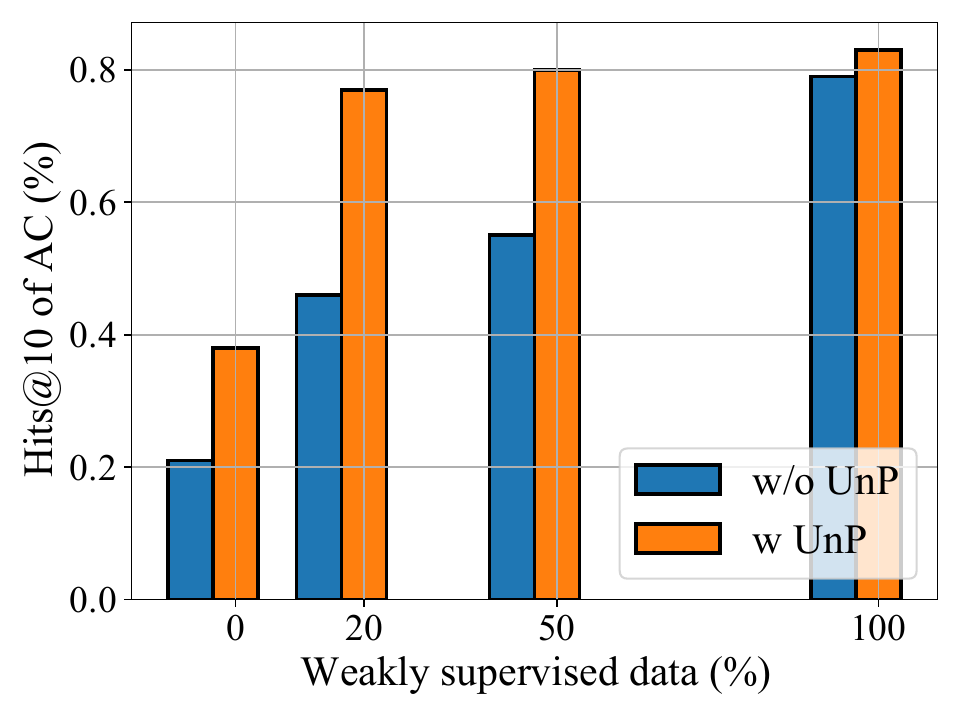}
	}
	\caption{\label{fig:trainingretriever} Retriever's performance by pre-training in terms of Hits@10 of answer coverage rate (AC). UnP denotes unsupervised pre-training.}
\end{figure}

\vpara{Effect of Question Update, Path Ending, and Subgraph Merge.}
We investigate the effects of the strategies used in \model, including the question updating strategy (QU) which concatenates the original question with the partially expanded path at each time step, the path ending strategy (PE) which learns when to stop expanding the path, and the subgraph merging strategy (GM) which induces a subgraph from the top-$nK$ paths.

Table~\ref{tb:retrieverperformance} indicates that based on \model, Hits@1 drops 4.3-15.0\% when removing QU (\smodel w/o QU) and Hits@1 drops 2.1-18.5\% when changing PE to the fixed path length $T$ (\smodel w/o PE), where the optimal $T$ is set to 3 on both WebQSP and CWQ. 

 Table~\ref{tb:treemerge} shows that based on \model+NSM, the average subgraph size increases from 174 to 204, and Hits@1 of QA drops 0.1\% when removing the subgraph merging strategy (\model+NSM w/o GM) but directly taking the union of all the subgraphs from different topic entities to induce the subgraph. We only present the results on CWQ as most of the questions in WebQSP only contain one topic entity, which does not need the merge operation.

\ipara{Summary.} The above results verify the effectiveness of the devised QU, PE, and GM in \model.

\begin{table}[t]
\newcolumntype{?}{!{\vrule width 1pt}}
\newcolumntype{C}{>{\centering\arraybackslash}p{2em}}
\caption{
	\label{tb:treemerge} Performance of subgraph merging strategy (GM) on CWQ.  
}
\centering 
\renewcommand\arraystretch{1.0}
\scalebox{0.8}{
\begin{tabular}{@{}l?cc@{}}
\toprule
Model & Subgraph size  & Hits@1 of QA (\%) \\ \midrule
\model+NSM             & 174    & 50.2  \\
\model+NSM w/o GM      & 204    & 50.1 \\
\bottomrule
\end{tabular}
}
\end{table}

\subsection{Training Strategy Evaluation}
\vpara{Effect of Pre-training.}
We investigate the effects of the weakly supervised and the unsupervised pre-training on the \model.
Table~\ref{tb:retrieverperformance} shows the performance of the supervised training (\smodel w SuperT) and the weakly supervised pre-training (\model), which indicates that \smodel is comparable with \smodel w SuperT when retrieving top-10 paths. Because a single ground-truth path between a topic entity and an answer is provided by WebQSP, which might omit the situation when multiple ground truth paths can be found. 
In view of this, the weakly supervised way that retrieves multiple shortest paths as the ground truth can provide richer supervision signals. We ignore the supervised training in CWQ because the ground truth paths are not explicitly given in the dataset.

We further vary the proportion of the weakly supervised data in \{0\%, 20\%, 50\%, 100\%\}, and present the corresponding answer coverage rate of the subgraph induced by top-10 paths (\emph{i.e.} Hits@10) in Figure~\ref{fig:trainingretriever}. Note 0\% means the RoBERTa used in \smodel don't have any fine-tuning. The performance shows a consistent growth with the weakly generated data size, which demonstrates its positive effect.

Before the weakly supervised pre-training, we create 100,000 pseudo instances for unsupervised pre-training (Cf. Section~\ref{sec:training} for details). 
The results presented by the orange bars show that unsupervised pre-training can significantly improve the original \smodel (0\% weakly supervised data) by about 20\% Hits@1. However, with the increase of the weakly-supervised data, adding unsupervised pre-training does not take better effect.

\ipara{Summary.} The above results show the effectiveness of the weakly supervised pre-training. Meanwhile, the unsupervised strategy can be an alternative choice when the QA pairs are scarce.

\vpara{Effect of End-to-End Fine-tuning.}
Table~\ref{tb:retrieverperformance} shows both \model+NSM w E2E and \model+GN w E2E improve 2-10.6\% Hits@1 of retrieval based on \model.
Table~\ref{tb:overall} shows  
\model+NSM w E2E improves 0.6\% Hits@1 of QA based on \model+NSM on WebQSP, and \model+GRAFT-Net w E2E improves 1.5-2.5\% Hits@1 of QA based on \model+GRAFT-Net. Although \model+NSM w E2E underperforms \model+NSM on CWQ, we suggest to reason on the top-1 retrieved results, which are much better than those before fine-tuning .      

\ipara{Summary.} The above results indicate that the answer likelihood estimated by the reasoner provides positive feedback for fine-tuning the retriever. 
With the improvement of the retriever, the reasoner can be also enhanced by the updated subgraphs.

\section{Conclusion}
We propose a subgraph retriever (\model) decoupled from the subsequent reasoner for KBQA. \smodel is devised as an efficient dual-encoder that can update the question when expanding the path as well as determining the stop of the expansion. The experimental results on two well-studied benchmarks show \smodel takes effect in improving the QA performance if injecting it before a subgraph-oriented reasoner. \smodel equipped with NSM creates new SOTA results for embedding-based KBQA methods if learning \smodel by weakly supervised pre-training as well as end-to-end fine-tuning.

\section*{Acknowledgments}
This work is supported by National Natural Science Foundation of China (62076245, 62072460, 62172424); National Key Research \& Develop Plan(2018YFB1004401); Beijing Natural Science Foundation (4212022); CCF-Tencent Open Fund.

\bibliography{anthology}
\bibliographystyle{acl_natbib}

\appendix \label{appendix}
\section{Appendix}
\label{sec:appendix}

\subsection{Interpretability of Retrieved Paths.}
We present the top-$nK$ paths learned by the proposed \smodel for several questions in Table~\ref{tb:casestudy} on WebQSP and CWQ. Each path is denoted by its topic entity before the colon. A path denoted by * means it is the new path discovered by \smodel beyond the ground-truth path provided by WebQSP and CWQ. The paths can explain why an answer is inferred for a question.

\begin{table*}[t]
\newcolumntype{?}{!{\vrule width 1pt}}
\newcolumntype{A}{>{\centering\arraybackslash}p{25em}}
\newcolumntype{C}{>{\centering\arraybackslash}p{20em}}
\caption{
\label{tb:casestudy}  Paths learned by the proposed retriever \model. 
}
\centering 
\renewcommand\arraystretch{1.0}
\scalebox{0.83}{
\begin{tabular}{@{}c@{ }?l@{}}
\toprule
Question & Top-$nK$ ($K$=2, $n$ is \#topic entities) Retrieved Paths  \\
\midrule
\multirow {2}{*}{Which airport to fly into <Rome>?} 
& $p_1$ = Rome:(travel\_destination, transportation) \\
& $p_2$ = Rome:(nearby\_airports)* \\ \midrule
\multirow {2}{*}{\tabincell{c}{<Santo Domingo> is the capital of the country\\ with what currency?}} 
& $p_1$ = Santo Domingo:(capital\_of, currency\_used) \\
& $p_2$ = Santo Domingo:(containedby, currency\_used)*\\\midrule 
\multirow{2}{*}{\tabincell{c}{What is the home county of the person who said <``Few \\people 
 have the virtue to withstand the highest bidder?''>?}}
& $p_1$ = (quotation\_author, place\_of\_birth)\\
& $p_2$ = (quotation\_author, places\_lived)*\\ \midrule
\multirow{2}{*}{What is the most recent movie directed by <Angelina Jolie>?}
& $p_1$ = Angelina Jolie:(director\_of\_film, release\_date)\\ 
& $p_2$ = Angelina Jolie:(producer\_of\_film, release\_date)*\\ \midrule
\multirow{4}{*}{\tabincell{c}{What movie, written by <Katerine Fugate>,\\featured <Taylor Swift>?}}
& $p_1$ = Katerine Fugate:(film\_writer) \\ 
& $p_2$ = Katerine Fugate:(film\_story\_contributor)* \\ 
& $p_3$ = Taylor Swift:(actor\_film) \\
& $p_4$ = Taylor Swift:(nominations, nominated\_for)* \\ \midrule
\multirow{4}{*}{\tabincell{c}{What country bordering <Argentina> \\is in the <Brasilia Time Zone>?}} 
& $p_1$ = Argentina:(location\_adjoins)\\ 
& $p_2$ = Argentina:(import\_from)* \\
& $p_3$ = Brasilia Time Zone:(location\_time\_zone) \\
& $p_4$ = Brasilia Time Zone:(location\_used, contained\_by)* \\ \midrule
\multirow{2}{*}{Where was the artist that had <This Summer Tour> raised?} 
& $p_1$ = This Summer Tour:(music\_artist, place\_of\_birth) \\ 
& $p_2$ = This Summer Tour:(music\_artist, artist\_origin)* \\ \midrule
\multirow{2}{*}{What position did <Vince Lombardi> play?} 
& $p_1$ = Vince Lombardi:(basketball\_player\_position)\\ 
& $p_2$ = Vince Lombardi:(teams, team\_roster\_position)* \\ \midrule
\multirow{2}{*}{What city is <Acadia University> in?} 
& $p_1$ = Acadia University:(headquarters, mailing\_address)\\
& $p_2$ = Acadia University:(contained\_by)*\\ \midrule
\multirow{2}{*}{\tabincell{c}{What was the first book written by \\ the author of <``The Cricket on the Hearth''>?}} 
& $p_1$ = ``The Cricket on the Hearth'':(author, works\_written) \\ 
& $p_2$ = ``The Cricket on the Hearth'':(author, book\_published)* \\ \midrule
\multirow{2}{*}{Who married to the person who lived in <Downe, Kent>?} 
& $p_1$ = Downe, Kent:(person\_lived, spouse) \\ 
& $p_2$ = Downe, Kent:(person\_lived, children, parent)* \\ \midrule
\multirow{2}{*}{\tabincell{c}{Find the location of the film <Fan Chan>, \\ what language is spoken there?}}
& $p_1$ = Fan Chan:(film\_location, languages\_spoken)\\ 
& $p_2$ = Fan Chan:(film\_location, official\_language)* \\ \bottomrule
\end{tabular}
}
\end{table*}

\subsection{Training Algorithm.}
We present the whole training process in Algorithm~\ref{algo:training}, where we first pre-train the retriever, then train the reasoner based on the retrieved subgraph, and finally end-to-end fine-tune the retriever and the reasoner together.

\begin{algorithm}
    \caption{Training Algorithm\label{algo:training}}
    \KwIn{$G,\{(q,a)\}$}
    \KwOut{Learned parameters $\theta$ and $\phi$.}
     Pre-train the retriever by weakly supervised signals or unsupervised signals plus only 20\% weakly supervised signals;\\
     Train the reasoner on the retrieved subgraphs;\\
    \tcc{End-to-End training:}
    \While{not converge}{
         For each ($q$, $a$) pair, sample a subgraph $\mathcal{G}$ by current retriever;\\
         Update $\phi$ by optimizing the first term of Eq.~\eqref{eq:loss} on all the ($q$, $a$, $\mathcal{G}$) instances;\\
         Update $\theta$ by optimizing the second term of Eq.~\eqref{eq:loss} on all the ($q$, $a$, $\mathcal{G}$) instances;\\
    }
\end{algorithm}


\subsection{Experimental Implementation}
\label{sec:implementation}
We provide the training and inference details of all the experiments as below. 

\vpara{General Setting.}
We use RoBERTa-base in our paper~\cite{liu2019roberta}. The basic RoBERTa contains 12 layers, 768-d hidden size, and 12 attention heads, resulting in 110M parameters in total. 
On WebQSP and CWQ, the batch size for training both the retriever and the reasoner is set as 16 and 20 respectively. 

\vpara{Supervised Training.}
WebQSP provides the relation chains corresponding to each (question, answer) pair. 
For each question, we use each relation chain from each topic entity to the answer as the ground truth path. In this way, we obtain 3,098 (question, path) instances which can be decomposed into 5,394 (question, relation) instances in total for supervised training. The learning rate for supervised training is set as 5e-5. An epoch takes about 5 minutes and the loss function converges within 10 epochs on WebQSP/CWQ. 
We ignore supervised training on CWQ because the explicit paths are not provided.

\vpara{Weakly supervised Pre-training.}
For weakly supervised pre-training, the ground truth paths are unavailable. To create the pseudo paths, for each (question, answer) pair, we extract all the shortest paths between each topic entity and an answer. We create 16,000/150,000 (question, relation) instances in total for weakly supervised pre-training.
The learning rate for weakly supervised training is set as 5e-5.
An epoch takes about 5 minutes and the loss function converges within 10 epochs.

\vpara{Unsupervised Pre-training.}
From the NYT dataset~\cite{riedel2010modeling}, we create 100,000 (sentence, path) pseudo instances. 
The learning rate for unsupervised training is set as 5e-5.
An epoch takes about 5 minutes and the loss function converges within 10 epochs. 
The unsupervised pre-training is performed once and then \smodel can be adapted to various KBQA datasets.

\vpara{End-to-End Training.}
Before end-to-end training, the retriever needs to be warmed up by weakly supervised pre-training or unsupervised pre-training. The reasoner also needs to be warmed up by supervised training on the  (question, answer) pairs.
For training the NSM reasoner~\cite{He_2021}, an epoch with batch size 20 takes 55 seconds and the loss function converges within 80 epochs. 
The learning rate for warming up the reasoner is set as 1e-4.
For end-to-end training, the learning rate is set as 1e-5.

\subsection{Inference}
We retrieve the top 10 relevant relations at each step which results in 10 paths for each topic entity. The number 10 is determined at the pre-training stage by checking the inflection point of the answer coverage rate on the validation set. 
The average time of online inference including both the subgraph retrieving and the reasoning can be within 1 second. By comparison, GRAFT-Net and NSM which first retrieve the whole two-hop subgraph and then prune it by the PPR scores spend about 2 to 3 seconds or even 7 to 8 seconds for retrieving some dense subgraphs.

\hide{
\subsection{Reproducibility Checklist.}

\vpara{A clear description of the mathematical setting,
algorithm, and/or model:} This is provided in the main paper in Section~\ref{sec:retriever} and Section~\ref{sec:training}.

\vpara{A link to a downloadable source code, with
specification of all dependencies, including
external libraries (recommended for camera
ready, though welcome for initial submission):} Our code is clearly organized and is available at 
\url{https://github.com/RUCKBReasoning/SubgraphRetrievalKBQA}.

\vpara{A description of computing infrastructure
used:} We run experiments with one RTX 3090(24G) GPU on the server with 256G physical memory and 8T disk size.

\vpara{The exact number of training and evaluation runs:} We report the runtime of different training methods in Appendix~\ref{sec:implementation}.

\vpara{The number of parameters in each model:} We provide the number of parameters in Appendix~\ref{sec:implementation}.

\vpara{The bounds for each hyperparameters:} We mainly tune the learning rate $\in [1e^{-4}, 1e^{-5}]$.

\vpara{Details of train/validation/test splits, data statistics and download link:} The training /validation/test splits are shown in Table~\ref{tb:dataset}. WebQSP is available at: \url{https://www.microsoft.com/en-us/download/details.aspx?id=52763} and CWQ is available at: \url{https://www.tau-nlp.org/compwebq}. 

}

\end{document}